# Parametrising the Inhomogeneity Inducing Capacity of a Training Set, and its Impact on Supervised Learning[*]

Gargi Roy[†] and Dalia Chakrabarty[‡]

**Abstract.** We introduce parametrisation of that property of the available training dataset, that necessitates an inhomogeneous correlation structure for the function that is learnt as a model of the relationship between the pair of variables, observations of which comprise the considered training data. We refer to a parametrisation of this property of a given training set, as its "inhomogeneity parameter". It is easy to compute this parameter for small-to-large datasets, and we demonstrate such computation on multiple publicly-available datasets, while also demonstrating that conventional "non-stationarity" of data does not imply a non-zero inhomogeneity parameter of the dataset. We prove that - within the probabilistic Gaussian Process-based learning approach - a training set with a non-zero inhomogeneity parameter renders it imperative, that the process that is invoked to model the sought function, be non-stationary. Following the learning of a real-world multivariate function with such a Process, quality and reliability of predictions at test inputs, are demonstrated to be affected by the inhomogeneity parameter of the training data.

**Key words.** Correlation structure of a random function, Non-stationary stochastic processes, Probabilistic learning, Logic in artificial intelligence, Applications of Markov chains and discrete-time Markov processes

**MSC codes.** 62H20, 60G10, 68T05, 68T27, 60J20

**1. Introduction.** Mechanistic learning is currently undertaken in applications across diverse domains, including healthcare [5]; finance [18]; education [24]; environment [25]; cybersecurity [16]; manufacturing [4]; robotics [35]; amongst others. In particular, supervised learning is ubiquitously implemented, as it enables learning of the function that represents the relationship between two variables given a training dataset, in order to ultimately perform prediction of the output of this sought function at a test input. Such predictions have been undertaken widely: in connection to epidemiology [26]; mining [2]; energy storage and management systems [37, 30]; agriculture [1]; construction engineering [3], etc. Predictive modelling is also used in accuracy-critical domains (such as information security [34]; medicine [9]), where the learning is required to be robust, interpretable, and uncertainty-included, as well as enabled to acknowledge the noise that is present in the training dataset [3, 31, 33, 30].

In the general context of supervised learning, the output variable can be high-dimensional, such that the function that links the input and output variables, can be multivariate and high dimensional in general. For example, we might seek a tensor-valued multivariate function that outputs a tensor, at a given value of the input vector. When undertaking probabilistic learning of the relationship between and output and an input - with said relationship modelled as a random function treated as a sample function of an adequately-chosen stochastic process - we wish to retain generalisability of the adopted learning strategy across dimensions, and ensure that minimal constraints be imposed on the sought function as we model it with the stochastic process. Both these requisites lead us to consider the process to be a Gaussian Process (GP) [29, 12, 13]. Faced with real-world training sets, this generative GP is often non-stationary, which then imposes a non-stationary correlation structure on the sought function. In other words, the correlation between outputs realised at a given input pair, is then anticipated to vary with shift in input space.

[*]Submitted to the editors DATE.
 **Funding:** No funding.
[†]Department of Industrial Engineering & Management, Ben-Gurion University of Negev, David Ben Gurion Blvd 1, Be'er Sheva, Israel (roygar@post.bgu.ac.il)
[‡]Department of Mathematics, University of York, Heslington, York YO10 5DD, United Kingdom (dalia.chakrabarty@york.ac.uk)

Then a crucially pertinent question arises about the nature of the data that renders the sought function to be such, that it cannot be modelled with a stationary GP. In the literature, this has sometimes been termed as "non-stationarity of the data". Indeed, [15] developed a measure of non-stationarity of data - in the context of time series data. [22] has proposed a method to quantify the non-stationarity of a stochastic process using information theoretic techniques. [28] presented a dedicated measure of non-stationarity based on the idea of model misspecification, to test the performance of their proposed kernels when applied to their test datasets, though the generalisability of this measure for generic datasets has not been discussed.

When the training set is such that it demands the learning of the sought inter-variable relationship as a function that can be generated only by a non-stationary process, we refer to the training data as capacitated to induce "inhomogeneities in the correlation structure" of the sought function. In this paper we introduce a parametrisation of this capacity of the data as the "inhomogeneity parameter of the data". We prove that such inhomogeneities in the training set implies that the sought function needs to be generated by a non-stationary process. Higher the value of this inhomogeneity parameter, more non-stationary the process needs to be, in order to render predictions at test inputs correct. Our interest here is not so much in the learning with non-stationary generative processes, but establishment of a measure of a property of training sets, that drives the sought functions to manifest a non-stationary correlation structure.

We present experiments with real data sets - marked by scalar and vector-valued input variables - to demonstrate that higher values of the inhomogeneity parameter leads to comparatively greater difficulty with the learning of an inter-variable function, leading to bigger errors in prediction of outputs at test inputs. In light of this, we appreciate that knowing the inhomogeneity parameter of the given training set before undertaking the function learning, will allow selection of appropriate learning model.

We introduce a parametrisation of the inhomogeneity in the correlation structure of data in Section 2, and exemplify this with multiple real-world datasets in Section 2.5, while clarifying that non-stationarity of data does not necessarily imply such a non-zero value of this inhomogeneity parameter (in Section 2.4). In Section 3, we discuss the effect of non-zero value of such a parameter of a given training set, in enhancing errors made in predictions made at the test inputs in a test dataset. This is illustrated in Section 3.3. in the context of a real-world probabilistic supervised learning that precedes said predictions at four different test sets, that are diversely inhomogeneous. Finally this paper is concluded in Section 4.

**2. Inhomogeneity parameter of a dataset.** Here we aim to learn the random function $\boldsymbol{f}(\cdot)$, where $\boldsymbol{Y} = \boldsymbol{f}(\boldsymbol{x})$, with the (generally) $k$-th order tensor-valued output $\boldsymbol{Y} \in \mathcal{Y} \subseteq \mathbb{R}^{m_1 \times m_2 \times \ldots \times m_k}$ and the (generally) vector-valued input $\boldsymbol{X} \in \mathcal{X} \subseteq \mathbb{R}^d$. We learn $\boldsymbol{f}(\cdot)$ using the $N$-sized training set $\mathbf{D} = \{(\boldsymbol{x}_i, \boldsymbol{y}_i)\}_{i=1}^N$, where $\boldsymbol{y}_i$ is the value of $\boldsymbol{Y}$ realised at $\boldsymbol{X} = \boldsymbol{x}_i$. We will model the sought random function as a sample function of a stochastic process, which we choose to be a Gaussian Process - that we will refer to as $\mathcal{GP}$. Let $\mathcal{GP}$ have the mean function $\boldsymbol{\mu}(\cdot)$ and covariance function $\boldsymbol{K}(\cdot, \cdot)$. As stated in the introductory section, justification for the choice of the process as a Gaussian Process emerges from the low level of constraints that such a process offers to the sample function; generalisability across dimensions; and ease of computation. Then $\boldsymbol{f}(\cdot) \sim \mathcal{GP}(\boldsymbol{\mu}(\cdot), \boldsymbol{K}(\cdot, \cdot))$.

We want to define - and parametrise - inhomogeneities of the training data $\mathbf{D}$, where such a property implies that the correlation structure of the function $\boldsymbol{f}(\cdot)$ that is learnt given this data $\mathbf{D}$, is not homogeneous across the input space, i.e. $\exists i \neq i'$ and $j \neq j'$, $(i, j, i', j' \in \{1, \ldots, n\})$, s.t. $corr[\boldsymbol{Y}_i, \boldsymbol{Y}_j] \neq corr[\boldsymbol{Y}_{i'}, \boldsymbol{Y}_{j'}]$, with $|\boldsymbol{x}_i - \boldsymbol{x}_j| = |\boldsymbol{x}'_i - \boldsymbol{x}'_j|$, where $|\cdot|$ denotes a distance function in $\mathcal{X}$. Here, $\boldsymbol{Y}_m = \boldsymbol{f}(\boldsymbol{x}_m)$ for $m = i, j, i', j'$. Such an inhomogeneous correlation structure of the function $\boldsymbol{f}(\cdot)$ implies that the process that generates this function - namely, $\mathcal{GP}$ - is non-stationary. Below, we employ this intuition to define the parameter that measures the inhomogeneity of the data variable $\mathcal{D}$, one realisation of which is $\mathbf{D}$.

*Definition* 2.1. *Dataset* $\mathbf{D}' := \{(i, \boldsymbol{y}_i)\}_{i=1}^N$ *is referred to as the "reduced data" where $\boldsymbol{y}_i$ is the realisation of the variable $\boldsymbol{Y}_i = \boldsymbol{f}(\boldsymbol{x}_i)$, with $\boldsymbol{x}_i$ the $i$-th design point in the training data* $\mathbf{D} = \{(\boldsymbol{x}_i, \boldsymbol{y}_i)\}_{i=1}^N$, $i = 1, \ldots, N$.

*Remark* 2.2. The training set $\mathbf{D}$ comprises discrete pairs of values of the input and of the corresponding output variables. Then the pursuit of "inhomogeneities" in $\mathbf{D}$ appears to sound ambiguous; in fact, what we mean by this is the capacity for $\mathbf{D}$ to induce inhomogeneities in the correlation structure of a function that is learnt with $\mathbf{D}$. We render the quantification of this capacity well-defined, by computing the inhomogeneities induced by $\mathbf{D}$ in the correlation structure of the "benchmark function". This benchmark function is defined using nothing other than the reduced data $\mathbf{D}'$ that is defined using the training set $\mathbf{D}$.

For a scalar-valued sought function, this benchmark function can be uniquely defined as the piecewise linear function that joins the $i$-th data point of the reduced dataset $\mathbf{D}'$, to the $i+1$-th datum of $\mathbf{D}'$, i.e. joins the point $(i, \boldsymbol{y}_i)$ to the point $(i+1, \boldsymbol{y}_{i+1})$, with a mapping $\{1, \ldots, N\} \longrightarrow \mathcal{Y}$, where $Y \in \mathcal{Y}$, for $i \in \{1, 2, \ldots, N-1\}$. Thus, for a scalar-valued $Y$, such a piece that joins points $(i, \boldsymbol{y}_i)$ and $(i+1, \boldsymbol{y}_{i+1})$, is linear $\forall i \in \{1, \ldots, N-1\}$. However, for the general high-dimensional (tensor-valued) output $\boldsymbol{Y} \in \mathcal{Y}$, it is difficult to define this piecewise linear mapping $\{1, 2, \ldots, N-1\} \longrightarrow \mathcal{Y}$, [19], that is built with $N-1$ linear pieces, with each such piece respectively defined over each of the intervals $[1, 2), \ldots, [N-2, N-1), [N-1, N]$ that are defined using the ordered sequence of indices of the design points of the training set.

However, for our purposes of developing an inhomogeneity parameter of data $\mathbf{D}$ with such a benchmark function - that invokes nothing other than the data $\mathbf{D}$ for its construction - we need to only define the correlation between the (generally tensor-valued) output $\boldsymbol{Y}_i = \boldsymbol{f}(\boldsymbol{x}_i)$ and the output $\boldsymbol{Y}_j = \boldsymbol{f}(\boldsymbol{x}_j)$, for $\boldsymbol{x}_i, \boldsymbol{x}_j \in \{\boldsymbol{x}_1, \boldsymbol{x}_2, \ldots, \boldsymbol{x}_N\}$, i.e. for $\boldsymbol{x}_i$ and $\boldsymbol{x}_j$ that are design points in the training set $\mathbf{D}$. Below, we define this correlation $corr(\boldsymbol{Y}_i, \boldsymbol{Y}_j)$ using a distance function $d_{\mathcal{Y}}(\boldsymbol{Y}_i, \boldsymbol{Y}_j)$ in the space $\mathcal{Y}$ that hosts the tensor-valued outputs.

Thus, for our purposes of developing an inhomogeneity parameter of a given training dataset, we will only have to compute the distance between outputs of $\boldsymbol{f}(\cdot)$ realised at a pair of design points in $\mathbf{D}$. Now, irrespective of details of the construction of a piecewise-linear benchmark function $\boldsymbol{f}_0 : [1, N] \longrightarrow \mathcal{Y}$, given the reduced data $\mathbf{D}'$, its output at the $i$-th input, is $\boldsymbol{Y}_i$, $i = 1, \ldots, N$. Then the aforesaid distance $d_{\mathcal{Y}}(\boldsymbol{Y}_i, \boldsymbol{Y}_j)$ gives the correlation between outputs of $\boldsymbol{f}_0(\cdot)$, realised at inputs $i$ and $j$. In other words, such a distance function will provide correlations between outputs realised at input pairs of $\boldsymbol{f}(\cdot)$, as well as of $\boldsymbol{f}_0(\cdot)$.

Thus, we will compute the distances $d_{\mathcal{Y}}(\boldsymbol{Y}_1, \boldsymbol{Y}_2), \ldots, d_{\mathcal{Y}}(\boldsymbol{Y}_{N-1}, \boldsymbol{Y}_N)$. If these distances do not concur with each other - in a way that we define below - then we conclude that the correlations between realisations of the benchmark function at successive pairs of the design points of $\mathbf{D}'$ are unequal, i.e. the correlation structure of the benchmark function is inhomogeneous. We will use the proxy: "inhomogeneities in data $\mathbf{D}$" to denote that there exist inhomogeneities amongst correlations between realisations of the benchmark function - that is defined by nothing other than $\mathbf{D}$ - at successive pairs of design points. Then to compute the parameter $p_{\mathbf{D}} \in [0, 1]$ that measures "inhomogeneities in data $\mathbf{D}$", we will count the number of distance values - computed between outputs of the benchmark function at successive pairs of design points of $\mathbf{D}'$ - and normalise that number by $N - 1$.

### 2.1. A distance function in $\mathcal{Y}$.

**Definition 2.3.** *We define a distance function in space $\mathcal{Y} \subseteq \mathbb{R}^{(m_1 \times \ldots \times m_k)}$ as*

$$d_{\mathcal{Y}}(\boldsymbol{Y}_i, \boldsymbol{Y}_j) = \sqrt{-log(|corr(\boldsymbol{Y}_i, \boldsymbol{Y}_j)|)}, \quad \forall i, j \in \{1, \ldots, N\}.$$

That $d_{\mathcal{Y}}(\cdot, \cdot)$ is a distance in $\mathcal{Y}$, is stated in Theorem 2.4.

**Theorem 2.4.** *In a random function $\boldsymbol{f} : \mathcal{X} \longrightarrow \mathcal{Y}$ (that has an inhomogeneous correlation structure in general), the function*

$$d_{\mathcal{Y}}(\boldsymbol{Y}_i, \boldsymbol{Y}_j) = \sqrt{-log(|corr(\boldsymbol{Y}_i, \boldsymbol{Y}_j)|)}, \quad \forall i, j \in \{1, \ldots, N\},$$

*is a distance function in $\mathcal{Y}$, where $\boldsymbol{Y}_i = \boldsymbol{f}(\boldsymbol{x}_i)$, for $i \in \{1, \ldots, N\}$.*

*Proof.*
— For $\boldsymbol{Y}_i := \boldsymbol{f}(\boldsymbol{x}_i) \in \mathcal{Y}$, $i \in \{1, \ldots, N\}$, $d_{\mathcal{Y}}(\boldsymbol{Y}_i, \boldsymbol{Y}_j) = \sqrt{-log(|corr(\boldsymbol{Y}_i, \boldsymbol{Y}_j)|)}$ is symmetric in $\boldsymbol{Y}_i$ and $\boldsymbol{Y}_j$, $\forall i, j \in \{1, \ldots, N\}$.
— $d_{\mathcal{Y}}(\cdot, \cdot) \geq 0$.
— For $i = j$, $\sqrt{-log(|corr(\boldsymbol{Y}_i, \boldsymbol{Y}_j))|)} = 0$, i.e. for $i = j$, $d_{\mathcal{Y}}(Y_i, Y_j) = 0$ $\forall i = j$.
    Again, for any $i, j \in \{1, \ldots, N\}$, $d_{\mathcal{Y}}(Y_i, Y_j) = 0 \implies corr(\boldsymbol{Y}_i, \boldsymbol{Y}_j) = 1$. However, the function $\boldsymbol{f}(\cdot)$ is random $\implies \nexists i \in \{1, \ldots, N\}$ s.t. realisation of the random variable $\boldsymbol{f}(\boldsymbol{x}_i)$ implies deterministic information on the value of $\boldsymbol{f}(\boldsymbol{x}_j) \forall j \neq i; j \in \{1, \ldots, N\}$ in general. Hence, $|corr(\boldsymbol{Y}_i, \boldsymbol{Y}_j)| = |corr(\boldsymbol{f}(\boldsymbol{x}_i), \boldsymbol{f}(\boldsymbol{x}_j))| = 1 \implies i = j$.
— For inputs $\boldsymbol{x}_i \neq \boldsymbol{x}_j \neq \boldsymbol{x}_k$, where $i, j, k \in \{1, \ldots, N\}$, we assume

$$\sqrt{-log(|corr(\boldsymbol{Y}_i, \boldsymbol{Y}_j))|)} + \sqrt{-log(|corr(\boldsymbol{Y}_j, \boldsymbol{Y}_k)|)} < \sqrt{-log(|corr(\boldsymbol{Y}_i, \boldsymbol{Y}_k)|)}.$$

Since absolute correlation $\in [0, 1]$, both sides of this inequality are positive. Then squaring both sides, we get:

$$\left(\sqrt{-log(|corr(\boldsymbol{Y}_i, \boldsymbol{Y}_j)|)} + \sqrt{-log(|corr(\boldsymbol{Y}_j, \boldsymbol{Y}_k)|)}\right)^2 <$$
$$-log(|corr(\boldsymbol{Y}_i, \boldsymbol{Y}_k)|).$$
$$\text{or, } -log(|corr(\boldsymbol{Y}_i, \boldsymbol{Y}_j)|) - log(|corr(\boldsymbol{Y}_j, \boldsymbol{Y}_k)|) +$$
$$2\sqrt{log(|corr(\boldsymbol{Y}_i, \boldsymbol{Y}_j)|)log(|corr(\boldsymbol{Y}_j, \boldsymbol{Y}_k)|)} <$$
$$-log(|corr(\boldsymbol{Y}_i, \boldsymbol{Y}_k)|).$$
$$\text{or, } 2\sqrt{log(|corr(\boldsymbol{Y}_i, \boldsymbol{Y}_j)|)log(|corr(\boldsymbol{Y}_j, \boldsymbol{Y}_k)|)} <$$
$$(2.1) \qquad log\left(\frac{|corr(\boldsymbol{Y}_i, \boldsymbol{Y}_j)||corr(\boldsymbol{Y}_j, \boldsymbol{Y}_k)|}{|corr(\boldsymbol{Y}_i, \boldsymbol{Y}_k)|}\right).$$

Then, $|corr(\boldsymbol{Y}_i, \boldsymbol{Y}_j)||corr(\boldsymbol{Y}_j, \boldsymbol{Y}_k)| < |corr(\boldsymbol{Y}_i, \boldsymbol{Y}_k)|$ will imply that the RHS of Inequality 2.1 is negative. However, its LHS is positive. This case then presents a contradiction. It then follows that the initial assumption is false. (Indeed, for $\boldsymbol{x}_i = \boldsymbol{x}_j \neq \boldsymbol{x}_k$, the initial assumption is trivially false.)
Therefore

$$\sqrt{-log(|corr(\boldsymbol{Y}_i, \boldsymbol{Y}_j)|)} + \sqrt{-log(|corr(\boldsymbol{Y}_j, \boldsymbol{Y}_k)|)} \geq \sqrt{-log(|corr(\boldsymbol{Y}_i, \boldsymbol{Y}_k)|)},$$

$\forall i, j, k \in \{1, \ldots, N\}$.
Thus, $d_{\mathcal{Y}}(\cdot, \cdot)$ abides by the triangle inequality.
It then follows from the above, that $d_{\mathcal{Y}}(\cdot, \cdot)$ is a distance function in space $\mathcal{Y}$. ∎

**2.2. Inhomogeneity parameter.** The correlation structure of the benchmark function $\boldsymbol{f}_0(\cdot)$ is given by $corr(\boldsymbol{Y}_i, \boldsymbol{Y}_j), \forall i, j \in \{1, \ldots, N\}$, and then by Definition 2.3, this is equivalently given by $d_{\mathcal{Y}}(\boldsymbol{Y}_i, \boldsymbol{Y}_j), \forall i, j \in \{1, \ldots, N\}$. Then the capacity in dataset $\mathbf{D}$ to induce inhomogeneities in the correlation structure of the benchmark function - what we refer to above as the "inhomogeneity parameter of data $\mathbf{D}$" - is informed by $d_{\mathcal{Y}}(\boldsymbol{Y}_i, \boldsymbol{Y}_j), \forall i, j \in \{1, \ldots, N\}$. This is equivalently the inhomogeneity parameter of the reduced data $\mathbf{D}'$.

To compute this inhomogeneity parameter, we first compute the distance

$$L_i := d_{\mathcal{Y}}(\boldsymbol{Y}_i, \boldsymbol{Y}_{i+1}) = \sqrt{-log(|corr(\boldsymbol{Y}_i, \boldsymbol{Y}_{i+1})|)},$$

and check if there exists $\ell$ s.t. $L_\ell$ is "incompatible" with $L_i$, for $\ell \neq i$, for $i, \ell \in \{1, \ldots, N-1\}$. We define such incompatibility below. Below, we refer to the distance $L_i$ between the $i+1$-th and $i$-th outputs in the space of the outputs, as an "$L$-value".

**Definition 2.5.** *A tolerance band of $\delta_i \geq 0$ is imposed on $d_{\mathcal{Y}}(\boldsymbol{Y}_i, \boldsymbol{Y}_{i+1})$, $\forall i \in \{1, \ldots, N-1\}$. Then if $\exists \ell \in \{1, \ldots, N-1\}$ s.t.*

$$[d_{\mathcal{Y}}(\boldsymbol{Y}_\ell, \boldsymbol{Y}_{\ell+1}) - \delta_\ell, d_{\mathcal{Y}}(\boldsymbol{Y}_\ell, \boldsymbol{Y}_{\ell+1}) + \delta_\ell] \cap [d_{\mathcal{Y}}(\boldsymbol{Y}_i, \boldsymbol{Y}_{i+1}) - \delta_i, d_{\mathcal{Y}}(\boldsymbol{Y}_i, \boldsymbol{Y}_{i+1}) + \delta_i] = \phi,$$
$$\text{for } i \in \{1, \ldots, \ell-1, \ell+1, \ldots, N-1\},$$

then datum $(\ell, \boldsymbol{y}_\ell)$ in the reduced dataset $\mathbf{D}'$ is defined as incompatible with the other data points $\{(i, \boldsymbol{y}_i)\}_{i=1; i \neq \ell}^N$ in $\mathbf{D}'$.

Then by definition, the inhomogeneity parameter $p_\mathbf{D} \in [0,1]$ for any training set $\mathbf{D}$.

*Remark* 2.6. The tolerance band $\delta_i$ on $L_i := d_\mathcal{Y}(\boldsymbol{Y}_i, \boldsymbol{Y}_{i+1}) = \sqrt{-log(corr(\boldsymbol{Y}_i, \boldsymbol{Y}_{i+1}))}$, is chosen by the practitioner. Typically, we set $\delta_i = \beta d_\mathcal{Y}(\boldsymbol{Y}_i, \boldsymbol{Y}_{i+1})$, with $\beta$ chosen $\in (0,1)$.

Definition 2.7. *Given a dataset $\mathbf{D} = \{(\boldsymbol{x}_i, \boldsymbol{y}_i)\}_{i=1}^N$, we define the reduced dataset $\mathbf{D}' = \{(i, \boldsymbol{y}_i)\}_{i=1}^N$. Then the inhomogeneity parameter $p_\mathbf{D}$ of $\mathbf{D}$ is defined as the fraction of data points in $\mathbf{D}'$ that are incompatible. Thus,*

$$p_\mathbf{D} := \frac{m}{N-1}, \text{ where } m = |\boldsymbol{A}|,$$

*with* $\boldsymbol{A} := \{\ell : [L_\ell - \delta_\ell, L_\ell + \delta_\ell] \cap [L_i - \delta_i, L_i + \delta_i] = \phi; (\ell, \boldsymbol{y}_\ell) \in \mathbf{D}', (i, \boldsymbol{y}_i) \in \mathbf{D}', i \neq \ell\}$,
(2.2)

*where*

$$L_i := d_\mathcal{Y}(\boldsymbol{Y}_i, \boldsymbol{Y}_{i+1}) = \sqrt{-log(|corr(\boldsymbol{Y}_i, \boldsymbol{Y}_{i+1})|)}.$$

*In our work, we choose $\delta_i = \delta$, $\forall i \in \{1, \ldots, N\}$.*

*Remark* 2.8. If inhomogeneity parameter $p_\mathbf{D} > 0$ for given dataset $\mathbf{D}$, then we refer to $\mathbf{D}$ as "bearing inhomogeneities".

**2.3. Non-zero inhomogeneity parameter demands non-stationary GP for learning $\boldsymbol{f}(\cdot)$.**
We use the training dataset $\mathbf{D}$ to learn the sought function $\boldsymbol{f}(\cdot)$. If $\mathbf{D}$ bears a non-zero inhomogeneity parameter, then we need to invoke a non-stationary GP to learn $\boldsymbol{f}(\cdot)$. If on the other hand, $p_\mathbf{D} = 0$, then $\boldsymbol{f}(\cdot)$ can be learnt using a stationary GP.

Theorem 2.9. *For the k-th order tensor-valued random variable $Y = \boldsymbol{f}(\boldsymbol{x})$, the random function $\boldsymbol{f}(\cdot)$ is learnt by modelling it as a sample function of a GP, given training set $\mathbf{D}$. Then, for a chosen $\delta > 0$, $p_\mathbf{D} > 0$ implies that the invoked GP has to be non-stationary. However, if $p_\mathbf{D} = 0$, then $\boldsymbol{f}(\cdot)$ can be modelled as a sample function of a stationary GP.*

*Proof.* For training set $\mathbf{D} = \{(\boldsymbol{x}_i, \boldsymbol{y}_i)\}_{i=1}^N$ comprising standardised data on the output, modelling the sought (k-th order tensor-valued) function $\boldsymbol{f}(\cdot)$ with a (tensor-valued) GP implies that the joint density $f_{\boldsymbol{Y}_1, \ldots, \boldsymbol{Y}_N}(\boldsymbol{y}_1, \ldots, \boldsymbol{y}_N)$ of $\boldsymbol{Y}_1, \ldots, \boldsymbol{Y}_N$, is a $k+1$-th tensor Normal density with a zero mean and $k+1$ correlation matrices, where $\boldsymbol{Y}_i = \boldsymbol{f}(\boldsymbol{x}_i)$, $(i = 1, \ldots, N)$.

Of these correlation matrices, the $N \times N$-dimensional correlation matrix is $\boldsymbol{\Sigma}_N = [corr(\boldsymbol{Y}_i, \boldsymbol{Y}_j)]$ is kernel parametrised with the kernel function $K(\cdot, \cdot)$ s.t. $\boldsymbol{\Sigma}_N = [corr(\boldsymbol{Y}_i, \boldsymbol{Y}_j)] = [K(\boldsymbol{x}_i, \boldsymbol{x}_j)]$. Here $i, j \in \{1, \ldots, N\}$.

The $N$ design points in training set $\mathbf{D}$ are marked with the index $i = 1, \ldots, N$. Now, given $\delta > 0$, $p_\mathbf{D} = m/(N-1)$ implies

— $\exists m$ values of $q$ s.t. $L_q$ is incompatible with $L_1, \ldots, L_{q-1}, L_{q+1}, \ldots, L_{N-1}$, i.e. $[L_q - \delta, L_q + \delta] \cap [L_i - \delta, L_i + \delta] = \phi$ for $i \in \{1, \ldots, q-1, q+1, \ldots, N-1\}$, (see Definition 2.7).
— Since $L_j := d_\mathcal{Y}(\boldsymbol{Y}_j, \boldsymbol{Y}_{j+1}) = \sqrt{-log(corr(\boldsymbol{Y}_j, \boldsymbol{Y}_{j+1}))}$, for $j \in \{1, \ldots, N-1\}$, it follows that $corr(\boldsymbol{Y}_q, \boldsymbol{Y}_{q+1}) \neq corr(\boldsymbol{Y}_i, \boldsymbol{Y}_{i+1})$, $\forall i \in \{1, \ldots, q-1, q+1, \ldots, N-1\}$, for $q = 1, \ldots, m$. But $corr(\boldsymbol{Y}_q, \boldsymbol{Y}_{q+1}) \neq corr(\boldsymbol{Y}_i, \boldsymbol{Y}_{i+1})$ implies $K(\boldsymbol{x}_i, \boldsymbol{x}_{i+1}) \neq K(\boldsymbol{x}_q, \boldsymbol{x}_{q+1})$.

— This is true even if $\parallel \boldsymbol{x}_i - \boldsymbol{x}_{i+1} \parallel = \parallel \boldsymbol{x}_q - \boldsymbol{x}_{q+1} \parallel$, where $\parallel \cdot \parallel$ is a norm of $\boldsymbol{x}$.
Then kernel $K(\cdot, \cdot)$ does not depend on the inputs only via a distance between them, but depends on the inputs individually. In other words, the kernel that can parametrise $\boldsymbol{\Sigma}_N$, is a non-stationary kernel, i.e. the generative tensor-valued GP is non-stationary.

So to summarise, $p_{\mathbf{D}} > 0 \implies$ tensor-valued GP that is used to model the tensor-valued sought function $\boldsymbol{f}(\cdot)$, is non-stationary.

Hence, a stationary GP can model the function if $p_{\mathbf{D}} \not> 0$, i.e. if $p_{\mathbf{D}} = 0$. ∎

**2.4. Non-stationarity of data D $\not\Rightarrow p_{\mathbf{D}} \geq 0$.** The testing of time series data for non-stationarity is commonly undertaken. For example, the Augmented Dickey Fuller test (ADF), tests for the null that the considered time series has a unit root, implying non-stationarity of this time series, (against the alternative that the time series at hand is stationary) [17, 27, 15, 8, 6]. Below, we provide an example to demonstrate that even if a dataset is non-stationary - as tested using ADF - the inhomogeneity parameter of this data can be 0.

*Remark* 2.10. The example below illustrates the assertion that the parametrisation of inhomogeneity of the data **D** via $p_{\mathbf{D}}$, does not inform on the pattern of variation in the values of the output across input values, but refers to that property of the data **D** that imparts a necessarily inhomogeneous correlation structure to any function, points on which include data points of **D**.

Example 1. *Consider the synthetic noisy nonstationary time series data $\mathbf{D}_{example}$ that is presented in the left panel of Figure 1. This dataset has been generated to comprise 52224 data points. We adduce evidence towards the non-stationarity of this time series by testing with ADF. The resulting test statistic is -1.228404 and the p-value is 0.661259. This then clearly states that the null cannot be rejected at significance levels of 1%, 5%, or 10%. (We recall that the critical values at these significance levels are respectively -3.430, -2.862, and -2.567 for the undertaken test).*

*However, the computed $p_{\mathbf{D}_{example}}$ of this dataset is nearly zero, (0.000038 to be precise), where the number m of "incompatible" L-values - out of the 52224-1=52223 points - is 2; (see Definitions 2.5 and 2.7). The tolerance band δ constructed around the value of the i-th output in this synthetic time series dataset, is chosen to be the maximal noise used in the data construction, at the i-th input, $\forall i \in \{1, \ldots, 52223\}$. The right panel of Figure 1 depicts the computed the L-values, plotted against the index of the data points in $\mathbf{D}_{example}$.*

*Indeed, the visualisation of the time series indicates its non-stationarity, and this example dataset illustrates that even when a data is not stationary, "inhomogeneity in the correlation structure of a data" can be (almost) 0.*

*Remark* 2.11. Only when the available training dataset **D** bears a non-zero inhomogeneity parameter - as parametrised by $p_{\mathbf{D}}$ - do we need to invoke a non-stationary $\mathcal{GP}$ to learn the function that is modelled as a sample function of this $\mathcal{GP}$, using this training set. It is such anticipated inhomogeneity in the correlation of the sought function and not the state of non-stationarity of the given data, that is relevant in our decision on whether a stationary $\mathcal{GP}$ can suffice for the learning exercise at hand, or not. Additionally, identification of non-stationarity of a sought high-dimensional data is difficult. However, the parameter $p_{\mathbf{D}}$ can be computed of such a dataset.

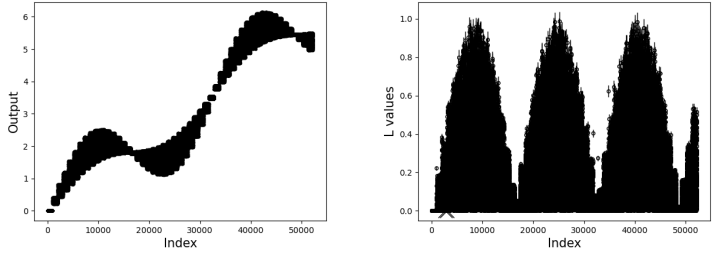

**Figure 1.** *Left: figure displaying the synthetic noisy time series data $\mathbf{D}_{example}$ discussed in Example 1. Right: L-values computed for each neighbouring pairs in this dataset, plotted against the index of the input values in this dataset.*

**2.5. Illustration of computation of the $p_{\mathbf{D}}$ parameter for real-world datasets.** We illustrate the computation of the inhomogeneity parameter on multiple real datasets that we refer to as: $\mathbf{D}_1$, $\mathbf{D}_2$, $\mathbf{D}_{2'}$, $\mathbf{D}_3$, $\mathbf{D}_4$ and $\mathbf{D}_5$. In said computation of $p_{\mathbf{D}}$, we have used $\delta = \delta_i = 0.05$, $\forall i \in \{1, \ldots, N-1\}$, for an $N$-sized dataset.

1. **Dataset $\mathbf{D}_1$:** this is an 184-sized subset of the Brent crude oil price data that is publicly available at https://datasource.kapsarc.org. This data comprises oil prices for the time interval ranging from the 5th of January, 2018, to the 29th of December, 2021, though the prices are not reported daily. The top left-most panel in Figure 2 presents the reduced version - which we refer to as $\mathbf{D}'_1$ - of the training set $\mathbf{D}_1$. The bottom left-most panel of Figure 2 presents the $L$-values. The inhomogeneity parameter $p_{\mathbf{D}_1}$ is 0.0326 with $m=6$. The ADF testing of the null that data $\mathbf{D}_1$ has a unit root results in a test statistic of -1.887949, (and a p-value: 0.337751). Then (recalling the critical values at significance of 1, 5 and 10% from the previous section) we cannot reject the null.

2. **Dataset $\mathbf{D}_2$:** this is a 200-sized subset of the power consumption data in Tetouan city, that is located in north Morocco [20, 32]. This data includes observations from the 1st of January, 2017, to the 31st of December of that year, recorded every 10 minutes. In fact, this recorded time series data includes values of five meteorological parameters - including temperature, humidity, wind speed, general diffuse flows and diffuse flows - and the corresponding value of the energy consumption in the three zones of this city, that the power distribution is divided into. The power consumption data across these three zones are added, and we refer to this sum as the "integrated power consumption" in this city. We will use this data later in Section 3 to learn the relationship between the integrated power consumption variable and the meteorological parameters, in order to predict the output power consumption, at test values of the input parameters. The second panel from the left in the top row of Figure 2 presents the plot of integrated power consumption against the index of the data points of this multivariate dataset. The second panel from the left in the bottom row, plots the $L$-values for this dataset. The inhomogeneity parameter $p_{\mathbf{D}_2} = 0.0345$ with 7 mutually-incompatible $L$-values. Hence, it can be seen that the data $\mathbf{D}_2$ bears a comparatively higher inhomogeneity than data $\mathbf{D}_1$. Testing for non-stationarity of data $\mathbf{D}_2$ using the ADF test, results in a test statistic of 0.31, (and a p-value: 0.98), such that we cannot reject the null.

3. **Dataset $\mathbf{D}_{2'}$:** this 200-sized dataset is another subset of the same Tetouan city that is used to create $\mathbf{D}_2$, though, distinguished from $\mathbf{D}_2$, this data is comparatively less inhomogeneously correlated. The third panel from the left in the top row of Figure 2 presents this dataset. The third panel from the left in the bottom row, plots the $L$-values for this dataset against the index of the inputs. The inhomogeneity parameter $p_{\mathbf{D}_{2'}}$ has been computed to be 0.005 with $m = 1$. Testing for non-stationarity of data $\mathbf{D}_{2'}$ using the ADF test, results in a test statistic of $-0.86$, (and a p-value of 0.8) such that we again cannot reject the null.

4. **Dataset $\mathbf{D}_3$:** this 347-sized dataset comprises information on the moving average of median values of sale prices of residential properties with single to five bedrooms, sold within a given geographical region. This house price data is obtained from https://www.kaggle.com/datasets/htagholdings/property-sales and includes property sales data for the period of 2007-2019. The fourth panel from the left in the top row of Figure 2 presents this dataset. The fourth panel from the left in the bottom row, plots the $L$-values for this dataset. The inhomogeneity parameter $p_{\mathbf{D}_3}$ has been computed to be 0.0259. Testing for non-stationarity of data $\mathbf{D}_3$ using the ADF test, results in a test statistic of $-2.03$, (and a p-value: 0.27). So we cannot reject the null.

5. **Dataset $\mathbf{D}_4$:** this is the 66497-sized ComEd data available at https://www.kaggle.com/datasets/robikscube/hourly-energy-consumption?select=COMED_hourly.csv. This dataset consists of the estimated energy consumption in (mega Watts) measured against time. The fifth panel from the left in the top row of Figure 2 presents the plot of energy consumption against the index of the data points in this dataset. The fifth panel from the left in the bottom row, displays the $L$-values for this dataset. The inhomogeneity parameter $p_{\mathbf{D}_4}$ has been computed to be: $p_{\mathbf{D}_4} = 0$. The ADF test of $\mathbf{D}_4$ suggests a test statistic of -15.6, which implies that we can reject the null that $\mathbf{D}_4$ is a non-stationary dataset, at significance levels of $1, 5$ and $10\%$. Thus, $\mathbf{D}_4$ is a dataset that bears zero inhomogeneity, and is also significantly not non-stationary.

6. **Dataset $\mathbf{D}_5$:** this is the large (121273-sized) PJM Energy Consumption Data available at https://www.kaggle.com/datasets/robikscube/hourly-energy-consumption. This data includes values of the power consumed per hour in mega Watts, for over 10 years, measured by electric transmission systems that serve various states of the USA. The data is released by PJM Interconnection LLC which is a regional transmission organisation in the USA, The data originally comes from PJM's website, and the consumed power is in units of mega Watts. The sixth panel from the left in the top row of Figure 2 presents the hourly power consumption against the index of the data point, with the sixth panel from the left in the bottom row, displaying a plot of the $L$-values for this data. The inhomogeneity parameter $p_{\mathbf{D}_5}$ has been computed to be as 0. ADF testing of $\mathbf{D}_5$ suggests a test statistic of about $-16.5$, leading to us rejecting the null that $\mathbf{D}_5$ is non-stationary, s.t. our conclusions about the non-stationary and inhomogeneity status of $\mathbf{D}_5$ and $\mathbf{D}_4$ are similar.

**3. Illustration of effect of $p_{\mathbf{D}}$ & on prediction, following probabilistic learning using data $\mathbf{D}_2$ & $\mathbf{D}_{2'}$.** In this section we will illustrate the connection between the inhomogeneity parameter of training datasets and the difficulty in predictions given a test set, depending on whether the learning acknowledges non-stationarity of the generative process. It is the input-output relationship - modelled as a random function - that is generated by the considered stochastic process. We will show that higher the value of the inhomogeneity parameter of the training data, higher is the gap in quality of predictions made using non-stationary model, compared to that made using a stationary model. We undertake this illustration by first non-parametrically learning a multivariate function, using a non-stationary GP, and separately, using a stationary GP. Both learning exercises are undertaken using each of two training sets that are disparate in their inhomogeneity-inducing capacity, namely, $\mathbf{D}_2$ and $\mathbf{D}_{2'}$, which comprise 200 data points. It is to be recalled that these two data sets are the subsets of the same Tetouan city data. Here $\mathbf{D}_2$ is much more inhomogeneously correlated than $\mathbf{D}_{2'}$; in fact, $p_{\mathbf{D}_2}$ for $\mathbf{D}_2$ is 0.0345 and for $\mathbf{D}_{2'}$ the $p_{\mathbf{D}_{2'}}$ is 0.005 . The undertaken learning is of the function that outputs the variable $Y$ that represents the integrated power consumption, realised at a given value of the four-dimensional input vector $\boldsymbol{X} = (X_T, X_H, X_W, X_D)^T$, where $X_T$ represents temperature; $X_H$ represents humidity; $X_W$ wind speed; and $X_D$ diffuse flows. (We found the input "general diffuse flows" to be a transformation of diffuse flows, and so dropped this attribute in the Tetouan city data, from our considered list of inputs). Once the probabilistic learning of this multivariate function is undertaken using a GP and a training set, predictions at 50 test inputs is made, and the quality of these predictions are inter-compared. Given that we are performing probabilistic learning using GPs, our prediction at any test input is closed-form; fast; and the predicted expected output at this test input is accompanied with the predicted variance of this output.

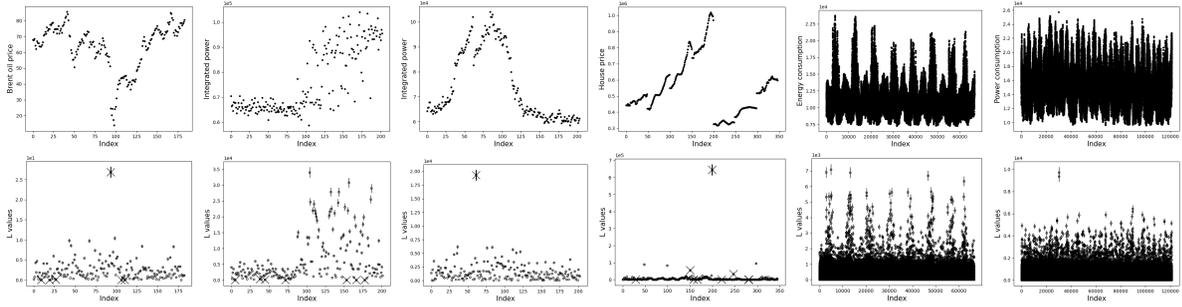

**Figure 2.** *Top row: output variable in six different publicly-available datasets or subsets of such datasets - introduced in the text as $\mathbf{D}_1, \mathbf{D}_2, \mathbf{D}_{2'}, \mathbf{D}_3, \mathbf{D}_4, \mathbf{D}_5$ - plotted respectively from left to right, against the index of the data points. Bottom row: computed L-values are plotted against the data index, for each of the datasets $\mathbf{D}_1, \mathbf{D}_2, \mathbf{D}_{2'}, \mathbf{D}_3, \mathbf{D}_4, \mathbf{D}_5$, from left to right. The tolerance interval $\delta_i$ on the i-th data point in any of these datasets, is chosen as 0.05, for all data points in each of these five datasets. This interval is depicted as an error bar (in black) that is superposed on each computed L-value. For a given dataset, incompatible L-values are marked with a cross.*

**3.1. Probabilistic learning using a non-stationary GP.** Let $Y \in \mathcal{Y}$ be a random variable that represents the integrated power consumed on any day in Tetouan city. Let $Y$ be associated with the following inputs: temperature ($X_T$); humidity ($X_H$); wind speed ($X_W$); diffuse flows ($X_D$), where we define $\boldsymbol{X} = (X_T, X_H, X_W, X_D)^T$. We will learn the unknown - and therefore modelled as random - function $f(\cdot)$, where

$$Y = f(x_T, x_H, x_W, x_D),$$

using the training data $\mathbf{D}_2 = \{(\boldsymbol{x}_i, y_i)\}_{i=1}^N$, where the $i$-th observed value of the input vector $\boldsymbol{X}$ is $\boldsymbol{x}_i = (x_T^{(i)}, x_H^{(i)}, x_W^{(i)}, x_D^{(i)})^T$, and $N = 200$. The training data $\mathbf{D}_{2'}$ has similar size and includes values of the same variables, but it is a distinct subset of the publicly-available Tetouan city dataset, compared to $\mathbf{D}_2$, s.t. $p_{\mathbf{D}_2} > p_{\mathbf{D}_{2'}}$.

Below, we discuss our learning of $f(\cdot)$ given training set $\mathbf{D}_2$. The same methodology underlines our learning using $\mathbf{D}_{2'}$. So we will not delineate that learning separately.

Let $y_i$ be the observed value of the output $Y_i \in \mathcal{Y} \subset \mathbb{R}$, realised at $\boldsymbol{X} = \boldsymbol{x}_i$, $\forall i \in \{1, \ldots, N\}$. In this learning exercise, we will standardise the data on the outputs, using the estimated mean and variance of the sample $\{y_1, \ldots, y_N\}$.

To learn $f(\cdot)$, we will model $f(\cdot)$ as a sample function $\tilde{f}(\cdot)$ of a GP that we denote $\mathcal{GP}$.

*Remark* 3.1. By definition of a Gaussian Process, any finite number of realisations of such a sample function $\tilde{f}(\cdot)$, are jointly distributed as a multivariate Normal density. So $Y_1, Y_2, \ldots, Y_N$ are jointly multivariate Normal, where each such output is realised at each of the $N$ design points $\boldsymbol{x}_1, \boldsymbol{x}_2, \ldots, \boldsymbol{x}_N$ that comprise the training set $\mathbf{D}_2$. Thus, the joint density $[Y_1, \ldots, Y_N] \equiv [\tilde{f}(\boldsymbol{x}_1), \tilde{f}(\boldsymbol{x}_2), \ldots, \tilde{f}(\boldsymbol{x}_N)]$ is the multivariate Normal density $MN(\boldsymbol{\mu}, \boldsymbol{\Sigma})$.

Since we have standardised the data on the outputs, we set $\boldsymbol{\mu}$ as a null vector, i.e. we use a zero-mean $\mathcal{GP}$ to model $f(\cdot)$, and $\boldsymbol{\Sigma}$ is reduced to a correlation matrix, the $ij$-th element of which is $corr(Y_i, Y_j)$, $\forall i, j \in \{1, \ldots, N\}$.

*Remark* 3.2. We kernel parametrise $\boldsymbol{\Sigma}$ with the kernel $K(\cdot, \cdot)$, s.t. $K(\boldsymbol{x}_i, \boldsymbol{x}_j) = corr(Y_1, Y_j)$ where $K(\cdot, \cdot)$ is a decreasing function of the difference between $\boldsymbol{x}_i$ and $\boldsymbol{x}_j$, and is also dependent on $\boldsymbol{x}_i$ and $\boldsymbol{x}_j$ individually. Thus, $K(\cdot, \cdot)$ is a non-stationary kernel, rendering $\mathcal{GP}$ a non-stationary process.

In Appendix A we discuss details of the modelling of the kernel as non-stationary.

As discussed in Appendix A, we invoke ergodicity of the system that generates the available training data $\mathbf{D}_2$, to model the $m$-th hyperparameter $\ell_m$ of the kernel $K(\cdot, \cdot)$ as $g_m(t)$, at the $t$-th post-burnin iteration of the undertaken MCMC inference used to perform the learning. Thus, in this iteration, $\ell_m = g_m(t)$, where $g_m(\cdot)$ is modelled as a sample function of a stationary Gaussian Process $\mathcal{GP}_m$, $\forall m \in \{T, H, W, D\}$. In this application, the input to the sought function $f(\cdot)$ is 4-dimensional, and the non-stationary kernel structure models the single kernel length scale hyperparameter along the $m$-th direction of the input space as a random function that is itself generated from a new GP (proven to be stationary) that we call $\mathcal{GP}_m$ s.t. the covariance function of $\mathcal{GP}_m$ is chosen to be parametrised with an SQE kernel bearing a single kernel hyperparameter $\delta_m$.

As stated in Appendix A, parameters of the correlation function of $\mathcal{GP}_m$ are learnt using a training set that we generate, and refer to as a "lookback data" comprising $T_L$ pairs of: (index of one of the previous $T_L$ iterations, the $\ell_m$ value that was current in that iteration of the MCMC chain).

In the first $N_b + T_L$ iterations of the MCMC - conducted with $N_{iter}$ iterations and burnin at $N_b$ - we will update the kernel hyperparameters $\ell_T, \ell_H, \ell_W, , \ell_D$ given the data $\mathbf{D}_2$, using

Metropolis Hastings, and the multivariate Normal likelihood with a null vector as the mean $\boldsymbol{\mu}$, i.e. $\boldsymbol{\mu} = \boldsymbol{0}$. Also, the correlation matrix of this multivariate Normal density is then $\boldsymbol{\Sigma}$, s.t. the likelihood is:

$$(3.1) \qquad \mathcal{L}(\theta_1, \ldots, \theta_H; \mathbf{D}_2) = \frac{1}{\sqrt{(2\pi)^N |\boldsymbol{\Sigma}|}} \exp\left(-\frac{1}{2}(\boldsymbol{y} - \boldsymbol{\mu})^T \boldsymbol{\Sigma}^{-1}(\boldsymbol{y} - \boldsymbol{\mu})\right).$$

Here $\boldsymbol{\Sigma}$ is kernel parametrised using an SQE kernel with hyperparameters $\ell_T, \ell_H, \ell_W, \ell_D$ that are modelled as unknown constants. Vague Normal priors with large variances are used on $\ell_m$, where such priors are centred at the chosen seeds of the hyperparameter values. We use the seed values of 0.1, 8.45, 0.02 and 0.015 for $\ell_T, \ell_H, \ell_W, \ell_D$ respectively, given the scale of values of the respective input variables in the data.

For iterations with index $t > N_b + T_L$, we use Metropolis-within-Gibbs to first update $\ell_T, \ell_H, \ell_W, \ell_D$ given data $\mathbf{D}_2$, using the same likelihood and priors as used during iterations with index $t \leq N_b + N_L$. Then we collate the values $l_{t-T_L}^{(m)}, \ldots, l_{t-1}^{(m)}$ of $\ell_m$ in the previous $T_L$ iterations into the lookback data that is the dynamic training set $\mathbf{D}_t^{(m)} := \{(c, l_c^{(m)})\}_{c=t-T_L}^{t-1}$. We recall that $m \in \{T, H, W, D\}$.

Using $\mathbf{D}_t^{(m)}$, we update the length scale hyperparameter $\delta_m$ of the stationary SQE kernel that parametrises the correlation matrix $\boldsymbol{\Psi}_m$ of the multivariate Normal density that is the joint of the $T_L$ random variables, $g_m(t - T_L), g_m(t - T_L + 1), \ldots, g_m(t - T_L + T_L - 1)$. Then ultimately, we predict the expectation of the variable $g_m(t - T_L + T_L)$, in a closed-form way, as is allowed by the Gaussian Process-based modelling of $g_m(\cdot)$.

This closed-form prediction of $\mathbb{E}[g_m(t - T_L + T_L)]$ is then treated as the current value of $\ell_m$ in the $t$-th iteration. Additionally, $\mathbf{D}_{t+1}^{(m)}$ is formulated with the last data point $(t + 1, \mathbb{E}[g_m(t - T_L + T_L)])$, while the first data point $(t - T_L, g_m(t - T_L))$ of $\mathbf{D}_t^{(m)}$ is dropped. Then in the first block of the $t+1$-th iteration, the updating of $\ell_m$ is performed using data $\mathbf{D}_2$, with $\mathbb{E}[g_m(t)]$ as the current value of $\ell_m$. However, in the second block of this $t+1$-th iteration, $\delta_m$ is updated using freshly updated lookback data $\mathbf{D}_{t+1}^{(m)}$. The above learning strategy, holds for $m \in \{T, H, W, D\}$.

Thus, learning the multivariate function $f(\cdot)$, given the dataset $\mathbf{D}_2$, reduces to the learning of the kernel hyperparameters $\ell_T, \ell_H, \ell_W, \ell_D, \delta_T, \delta_H, \delta_W, \delta_D$. As stated at the beginning of this subsection, we have similarly - and separately - undertaken the learning of $f(\cdot)$ as a sample function of a non-stationary GP, using dataset $\mathbf{D}_{2'}$. In Table 1, we present the mean values and the 95% HPD credible regions on values of the $\ell_m$ and $\delta_m$ parameters, learnt given the two datasets $\mathbf{D}_2$ and $\mathbf{D}_{2'}$, for $m \in \{T, H, W, D\}$. Appendix A includes the traces of these hyperparameters, as we made Metropolis-within-Gibbs based inference on these, given the two datasets.

**3.2. Probabilistic learning using a stationary GP.** In this subsection we discuss the learning of $f(\cdot)$ as a sample function of a stationary GP that we denote $\mathcal{GP}$. Thus, learning the multivariate function $f(\cdot)$ given $\mathbf{D}_2$ and $\mathbf{D}_{2'}$, reduces to the learning of the length scale kernel hyperparameters $\ell_T, \ell_H, \ell_W, \ell_D$ along the four directions of input space, where we choose to work with an SQE kernel. The aim is to compare the quality of predictions made after learning $f(\cdot)$ with a non-stationary GP, and predictions made subsequent to learning $f(\cdot)$ with a stationary GP - using a given training set. The mean values and the 95% HPD credible regions on learnt of these kernel hyperparameters, are presented in Table 2. In Appendix B the learning of these models are included.

|  | $\ell_T$ | $\ell_H$ | $\ell_W$ | $\ell_D$ | $\delta_T$ | $\delta_H$ | $\delta_W$ | $\delta_D$ |
|---|---|---|---|---|---|---|---|---|
| 95% HPD $\|\mathbf{D}_2$ | [2.07,2.09] | [15.9,16] | [0.0029,0.0032] | [0.0176,0.0187] | [0.16,0.24] | [0.36,0.6] | [0.81,0.93] | [0.29,0.48] |
| Mean $\|\mathbf{D}_2$ | 2.079 | 15.99 | 0.00305 | 0.0182 | 0.198 | 0.48 | 0.872 | 0.389 |
| 95% HPD $\|\mathbf{D}_{2'}$ | [0.78,0.793] | [8.68,8.7] | [0.0025,0.0028] | [0.039995,0.039998] | [0.38,0.42] | [0.21,0.52] | [0.73,0.92] | [0.18,0.22] |
| Mean $\|\mathbf{D}_{2'}$ | 0.79 | 8.69 | 0.0026 | 0.039997 | 0.39 | 0.36 | 0.82 | 0.2 |

**Table 1**

*Table displays the expected values and the learnt 95% Highest Probability Density credible region of the length scale hyperparameters $\ell_T, \ell_H, \ell_W, \ell_D$ of the non-stationary kernel that parametrises the correlation function of the non-stationary process $\mathcal{GP}$, which generates the sought multivariate function $f(\cdot)$. Results of learning undertaken with $\mathbf{D}_2$ are presented at the top and those with $\mathbf{D}_{2'}$ are in the bottom row. The table also displays values of hyperparameter $\delta_m$ of the kernel that parametrises the inner GP $\mathcal{GP}_m$ that generates the random function that outputs $\ell_m$, $\forall m \in \{T, H, W, D\}$.*

|  | $\ell_T$ | $\ell_H$ | $\ell_W$ | $\ell_D$ |
|---|---|---|---|---|
| 95% HPD $\|\mathbf{D}_2$ | [9.69,9.88] | [28.8,29.19] | [0.00069,0.0014] | [0.0059,0.0097] |
| Mean $\|\mathbf{D}_2$ | 9.79 | 28.99 | 0.0011 | 0.0078 |
| 95% HPD $\|\mathbf{D}_{2'}$ | [0.18,0.22] | [8.38,8.8] | [0.0075,0.037] | [0.016,0.041] |
| Mean $\|\mathbf{D}_{2'}$ | 0.1998 | 8.59 | 0.0223 | 0.0283 |

**Table 2**

*The top row displays the learnt 95% Highest Probability Density credible regions and mean values of hyperparameters of kernels that parametrise the correlation functions of the stationary GP, given the training set $\mathbf{D}$. datasets $\mathbf{D}_2$. the bottom row displays the same obtained from the learning undertaken with training set $\mathbf{D}_{2'}$.*

**3.3. Predictive results on test inputs, following learning done using $\mathbf{D}_2$ and $\mathbf{D}_{2'}$ in stationary and non-stationary models.** Once we can specify the correlation function of the process that generates the sought function $f(\cdot)$, we can compute (in a closed-form way), the expectation and variance of the output $Y_{test}$ that is realised at a test input.

Test inputs in the dataset "Test-1" is built from the publicly-available Tetouan city power consumption data set such that no test data points has any overlap with either training dataset $\mathbf{D}_2$ or $\mathbf{D}_{2'}$ that we use in this work. True values of the outputs realised at these test inputs are then also known, s.t. the error in a prediction can be computed.

We parametrise the error in prediction using Root Mean Square Error (RMSE) as well as the "Compatibility parameter" ($C$), where we define $C$ as the fraction of the number of predictions - undertaken at $N_{test}$ number of test inputs - s.t. the true value $y_i$ of the output at the $i$-th test input, is included within the predicted interval $[\bar{y}_i - s_i, \bar{y}_i + s_i]$. Here, $i \in \{1, \ldots, N_{test}\}$; $\bar{y}_i$ is the predicted expectation of the output realised at this $i$-th test input; and $s_i^2$ is the predicted variance of the output realised at this input. Thus, $C := |\boldsymbol{M}|/N_{test}$, where set $\boldsymbol{M} := \{i : y_i \in [\bar{y}_i - s_i, \bar{y}_i + s_i]; i \in \{1, \ldots, N_{test}\}\}$. On the other hand RMSE $= \sqrt{\sum_{i=1}^{N_{test}}(y_i - \bar{y}_i)^2/N_{test}}$.

The RMSE and $C$ values computed using the predictions made at the test inputs included within "Test-1" are presented Table 3. Here, the predictions are made subsequent to learning done by treating the sought function $f(\cdot)$ as a sample function of a stationary or non-stationary GP, given $\mathbf{D}_2$ or $\mathbf{D}_{2'}$. Figure 3 shows the predictions at the 50 considered test inputs, with predictive uncertainty (of one predicted standard deviation about the predicted mean) marked as error bars. True values of the output at any test input is also depicted in the figures. Since the input variable is a four-dimensional vector $\mathbf{X}$, we depict the prediction against each component $(X_T, X_H, X_W, X_D)$ of $\mathbf{X}$ and also, against the index of the test datum. This depiction of the prediction at the test inputs is undertaken following the learning done with a non-stationary GP and training set $\mathbf{D}_2$ in the top row; a non-stationary GP and training set $\mathbf{D}_{2'}$ in the 2nd row from top; a stationary GP and training set $\mathbf{D}_2$ in the 2nd row from the bottom; and a stationary GP and training set $\mathbf{D}_{2'}$ in the bottom row. Results are tabulated in Table 3.

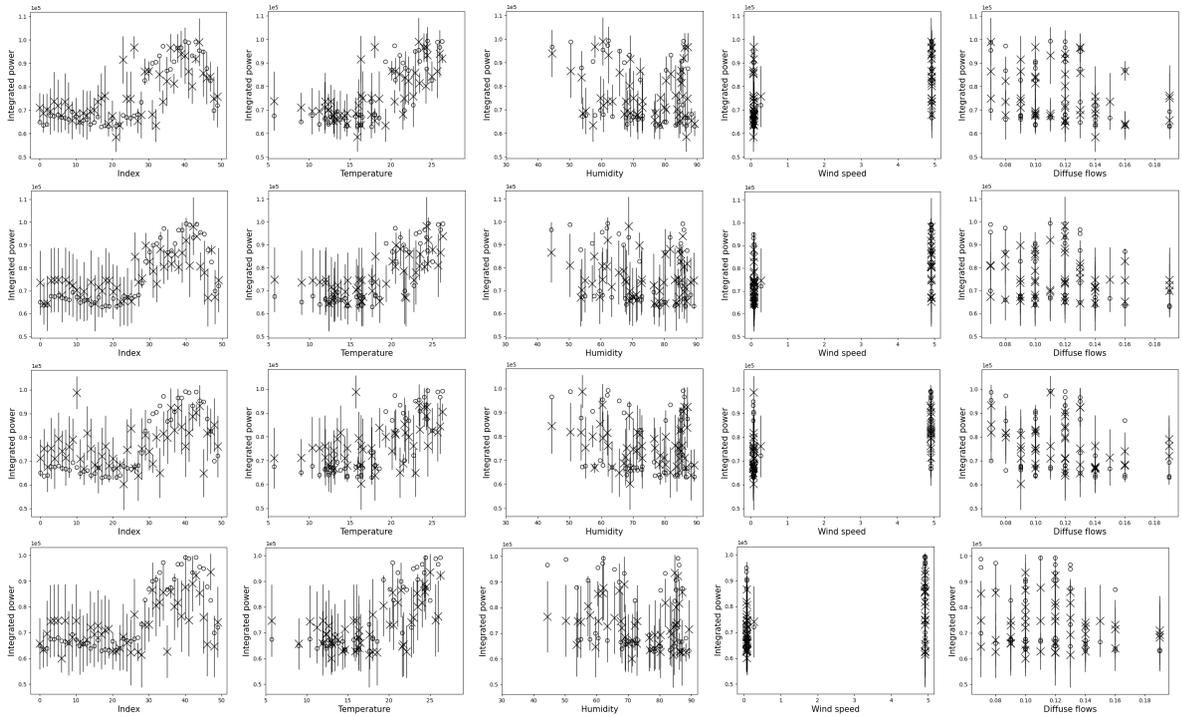

**Figure 3.** *In panels from left to right, we present plots of predicted mean values of the output (integrated power consumption), against the index of test data points; test values of $X_T$ (i.e. temperature); test values of $X_H$ (i.e. humidity); test values of $X_W$ (i.e. wind flows); and test values of $X_D$ (i.e. diffuse wind flows). Predictions made at 50 test inputs in "Test-1", following learning with a non-stationary GP, given $\mathbf{D}_2$, and $\mathbf{D}_{2'}$ are in the top and second-from-top rows respectively. True values of the output at each test input is known, and shown in open circles, while the predicted expectation is shown in crosses with the accompanying error bar representing one standard deviation predicted on the output realised at a test input. Predictions made with stationary GPs, given training sets $\mathbf{D}_2$ and $\mathbf{D}_{2'}$ are depicted in the third-from-top and bottom rows, respectively.*

### 3.4. Conclusions from the illustrated learning.

15| On "Test-1" | $\mathbf{D}_2$ ($p_{\mathbf{D}_2} = 0.0345$) | | $\mathbf{D}_{2'}$ ($p_{\mathbf{D}_{2'}} = 0.005$) | |
|---|---|---|---|---|
| | non-stationary | stationary | non-stationary | stationary |
| RMSE | 9875.2 | 11835.8 | 8759.9 | 9794.6 |
| C(%) | 74% | 64% | 74% | 70% |

**Table 3**
*Predictive performance by RMSE and Compatibility parameter (C) values, computed using predictions on expectation and variance of outputs realised at 50 test inputs that comprise "Test 1", in which we know the true output value at any test input. The predictions follow learning performed using both a stationary and non-stationary GP, given each of the two training sets $\mathbf{D}_2$ and $\mathbf{D}_{2'}$, that bear disparate inhomogeneity parameter values, $(p_{\mathbf{D}_2} > p_{\mathbf{D}_{2'}})$.*

*Remark* 3.3. From Table 3, we note that the quality of the prediction of function $f(\cdot)$ is better when we make its learning as a random sample function of a non-stationary GP, than a stationary GP, though the disparity in the prediction quality gets diluted with decreasing inhomogeneity-inducing capacity in the training dataset.

Thus, when the used training set is $\mathbf{D}_{2'}$ - which has a lower inhomogeneity-inducing capacity than $\mathbf{D}_2$ - the prediction quality is comparatively less different, if the learning were done using a stationary GP or a non-stationary GP, than when the used training set is $\mathbf{D}_2$. Here prediction quality, as assessed by the RMSE measure, imply a difference of 12% between predictions obtained after learning done with stationary and non-stationary models, when training is done with $\mathbf{D}_{2'}$. On the other hand, this RMSE difference is about 20% when training is done with $\mathbf{D}_2$. The disparity shows in the values of $C$ as well - 74% of the true output values lie within the $\pm 1$-standard deviation of the mean of the output predicted at the 50 test inputs, when the learning was done with a non-stationary GP, while this fraction was only 64% when the stationary GP was used given the more inhomogeneity-inducing data, though 70% compatibility between true and predicted values was noted when the less inhomogeneity-inducing training set was used in the stationary model.

*Remark* 3.4. Importantly, in our Bayesian setting, when we use the correct model - namely the non-stationary model - difference in the training set does not make much difference within the correctly identified uncertainties, i.e. when the Compatibility parameter $C$ is used to assess the quality of predictions.

**4. Conclusions.** The fundamental challenge in probabilistic supervised learning of a function that represents the relationship between an input and an output variable, is the learning of the correlation structure of this function. For a given input and output, learning said function given a training dataset, gets more difficult with increase in the inhomogeneity in the correlation structure of this function - after all, learning of a function is equivalent to learning its correlation, s.t. a correlation distribution (across the space of the input variable) that is further from being uniform, is harder to capture in the learning. Here by "inhomogeneity in the correlation structure" of this function, we imply that there is a difference between the correlation of a pair of outputs that are realised at two given input values, and the correlation of another output pair, realised elsewhere in input space.

The question that then arises, is if there is a property of the training dataset, that induces such inhomogeneity in the correlation structure of this function that we seek to learn, given this training set. Knowing the answer to this question is useful since we can then arrive at a data-driven decision on the requisites of the learning infrastructure, that will allow for reliable learning of the function, leading to accurate predictions. Here we provide a parametrisation $p_\mathbf{D}$ of such a property of the data $\mathbf{D}$ - we refer to it as the 'inhomogeneity parameter" of the data at hand. Such a parameter can be computed for data on outputs realised at design values of the input vector, even if the output variable is high-dimensional.

We prove that a function that is to be learnt with a training set $\mathbf{D}$, s.t. $p_\mathbf{D} > 0$, will manifest an inhomogeneous correlation structure, rendering it imperative to model such a sought function as a sample function drawn from a non-stationary Gaussian Process. Ideally, the eventual data-driven deliverable is expected to be a parametrisation of this non-stationarity of such a process, given the inhomogeneity of the training set. Such could be informed via a learning of the relationship between the variable that is $p_\mathbf{D}$ of training set $\mathbf{D}$, and a parametrisation of non-stationarity of the GP used to model the function that is learnt using this training data. This project is under consideration.

## Appendix A. Results of MCMC-based inference to facilitate non-parametric learning.

Traditional GP-based probabilistic learning of a sought function $f(\cdot)$, requires modelling $f(\cdot)$, (where $Y = f(x)$) with a sample function of the Gaussian Process $\mathcal{GP}$, leading to a multivariate Normal likelihood with the correlation matrix $\boldsymbol{\Sigma}$. This correlation matrix $\boldsymbol{\Sigma} = [corr(Y_i, Y_j)]$, $\forall i \in \{1, \ldots, N\}$. But we cannot learn every $N(N-1)/2$ distinct and unknown elements of $\boldsymbol{\Sigma}$ directly from MCMC, since $N(N-1)/2$ is a large number $= 19900$. To avoid this direct learning of the large number of correlations, we perform kernel parametrisation of the correlation function of $\mathcal{GP}$, using the kernel $K(\cdot, \cdot)$ s.t. $K(\boldsymbol{x}_i, \boldsymbol{x}_j) = corr(Y_1, Y_j)$, where $K(\cdot, \cdot)$ is a decreasing function of a difference between $\boldsymbol{x}_i$ and $\boldsymbol{x}_j$, and is also dependent on $\boldsymbol{x}_i$ and $\boldsymbol{x}_j$ individually, in the general case. In other words, the kernel $K(\cdot, \cdot)$ that is used, is a non-stationary kernel in general, implying that $\mathcal{GP}$ is a non-stationary process.

Now we discuss the non-stationary model that we use. The kernel $K(\cdot, \cdot)$ is manifestly non-stationary if its hyperparameters $(\ell_1, \ldots, \ell_H)$ are s.t. $\forall m \in \{1, 2, \ldots, H\}$, the variable $\ell_m$ attains a value at the input pair $\boldsymbol{x}_i, \boldsymbol{x}_j$, that is different from the value attained at the input pair $\boldsymbol{x}'_i, \boldsymbol{x}'_j$, for $(x_i, x_j) \neq (x'_i, x'_j)$. In other words, a distinct hyperparameter $\ell_m$ would need to be learnt at each input pair, where there are $N(N-1)/2 = 19900$ number of such input pairs in the $N = 200$-sized training sets $\mathbf{D}_2$ and $\mathbf{D}_{2'}$. Thus, in our model of non-stationarity, we would need to learn an infeasibly large number of kernel hyperparameters. So we need to avoid the direct learning of these $N(N-1)/2$ number of kernel hyperparameters, and we achieve this aim by resorting to a different modelling of the non-stationarity in the correlation function of the GP invoked to model the multivariate function $f(\cdot)$.

To develop a new model for the kernel $K(\cdot, \cdot)$, we recognise that the value of the kernel hyperparameter $\ell_m$ attained at the $u$-th pair of input values, will be different from that attained at the $v$-th pair of inputs, where, as we saw above, there are $N(N-1)/2$ distinct inputs to consider. We could then model $\ell_m$ as a random function $g_m(\cdot)$ - that takes as its input, the sample function of $\mathcal{GP}$ - s.t. $g_m(\cdot)$ is itself treated as a sample function of a distinct GP that we denote $\mathcal{GP}_m$. Then at the $u$-th pair of values of the variable $\boldsymbol{X}$, we could draw the $u$-th sample function from $\mathcal{GP}$, and at the $v$-th pair of values of $\boldsymbol{X}$, we could draw the $v$-th sample function from $\mathcal{GP}$, where $u, v \in \{1, \ldots, N(N-1)/2\}$. Each of such $N(N-1)/2$ number of sample functions sampled from $\mathcal{GP}$ will then need to be used as the input to $g_m(\cdot)$, to yield a distinct value of $\ell_m$, at each distinct pair of values of $\boldsymbol{X}$, where $g_m(\cdot)$ is itself a sample function drawn from $\mathcal{GP}_m$. However, learning the aforesaid random function $g(\cdot)$ - that takes a sample function as its input - is difficult. This problem is mitigated by assuming the system that produces the observable training data, is ergodic [23, 10].

*Remark* A.1. During any step of the learning, we realise that by the Ergodic theorem, [21, 7, 14], drawing $T_L$ generally distinct random functions - at each pair of values of $\boldsymbol{X}$, i.e. at each value of which, a value of $\ell_m$ is realised - is equivalent to instead drawing one random function at each of $T_L$ steps of the learning exercise, as long as $T_L$ is large. Here by "step of the learning" exercise, we imply an iteration of the Bayesian inference that is undertaken to perform said learning. For us, such inference is performed using Markov Chain Monte Carlo, or MCMC. The drawing of a sample function at each of $T_L$ iterations of MCMC is done after burnin of the Markov chain, i.e. after the chain has achieved equilibrium, s.t. sampling is done from processes that differ slightly - and in a trendless way - in their correlation functions.

Then we model $\ell_m = g_m(t)$, where $t$ is the value of the iteration index variable $T$ of the $N_{iter}$ steps-long undertaken MCMC, the burnin of which is at $T = N_b$. So for us, $t \in [N_b+1, N_{iter}]$. Here $g_m(\cdot)$ is a random function $g_m : [N_b+1, N_{iter}] \subset \mathbb{N} \longrightarrow \mathbb{R}$, $\forall m = 1, \ldots, H$. We learn $g_m(\cdot)$ at every iteration indexed by $T = t \in [N_b+1, N_{iter}]$, using the collated, $T_L$-sized, dynamic training data: $\mathbf{D}_t^{(m)} := \{(c, l_c^{(m)})\}_{c=t-T_L}^{t-1}$, where the value of $\ell_m$, current in the $c$-th iteration of the undertaken Markov chain is $l_c^{(m)}$.

It can be shown that such a function $g_m(\cdot)$ is continuous [36]. Then $g_m(\cdot)$ can be modelled using a stationary Gaussian Process that we denote $\mathcal{GP}_m$. This follows, since a stationary Gaussian Process permits either all continuous sample functions, or no sample function is bound within a finite interval, [11]. Now, sample functions that can model the essentially continuous function $g_m(\cdot)$, are all continuous. Therefore, all such invoked sample functions could be drawn from a stationary Gaussian Process, i.e. $g_m(\cdot)$ can be modelled as a sample function of the stationary process $\mathcal{GP}_m$. This holds for $m = 1, \ldots, H$. Thus, our learning of the unknown multivariate function $f(\cdot)$ invokes $H$ stationary Gaussian Processes, nested within an outer non-stationary process $\mathcal{GP}$.

Then at the $t$-th post-burnin iteration of the undertaken MCMC, by definition, the $T_L$ realisations $g_m(t-T_L), \ldots, g_m(t-1)$ of the random function $g_m(\cdot)$ are jointly distributed as a multivariate Normal density. Again, we will standardise the data on the outputs realised at the design times in the dynamic training set $\mathbf{D}_t^{(m)}$, s.t. the mean of this multivariate Normal is used as a null vector, and its $T_L \times T_L$-dimensional correlation matrix $\boldsymbol{\Psi}_m$ is kernel parametrised using a stationary SQE kernel: $\boldsymbol{\Psi} = [\exp(-(t_a - t_b)/\delta_m)]; t_a, t_b \in \{t - T_L, t - T_L, +1, \ldots, t - T_L + T_L - 1\}$. We then learn $\delta_m$ using the dynamic training data. We set $T_L \equiv T_L$ as the width of the lookback window, from the current time.

Figure 4 displays traces of the hyperparameters $(\ell_T, \ell_H, \ell_W, \ell_D)$ of the kernel that parametrises the correlation matrix of $\mathcal{GP}$, learnt given training set $\mathbf{D}_2$, as well as the traces of $\delta_T, \delta_H, \delta_W, \delta_D$ that are the hyperparameters of the stationary kernels of $\mathcal{GP}_T, \mathcal{GP}_H, \mathcal{GP}_W$, $\mathcal{GP}_D$. Again, Figure 6 displays the traces of the same parameters, learnt this time using the training set $\mathbf{D}_{2'}$. Here, $\ell_T$ is the length scale along the temperature input dimension; $\ell_H$ the humidity input dimension; $\ell_W$ to the wind flow dimension; and $\ell_D$ is the length scale along the diffuse flow dimension. Again, $\delta_i$ is the length scale of $\mathcal{GP}_m$; for $m \in \{T, H, W, D\}$. Figure 5 and 7 present traces of the logarithm of the posterior densities, given the training data $\mathbf{D}_2$ and $\mathbf{D}_{2'}$ respectively in the outer layer as well as given the dynamically-varying lookback data. The two MCMC chains given $\mathbf{D}_2$ and $\mathbf{D}_{2'}$, are run for the 90000 and 200000 iterations respectively, with the size of the lookback window $T_L$ set to 100. Normal proposal densities are used to propose values of the hyperparameters during inference. Normal priors are used on all unknown hyperparameters. Variances of these Normal priors, as well as the Normal proposals are chosen experimentally. We start the chains with large variances, and then tighten the priors as convergence is noted to show up.

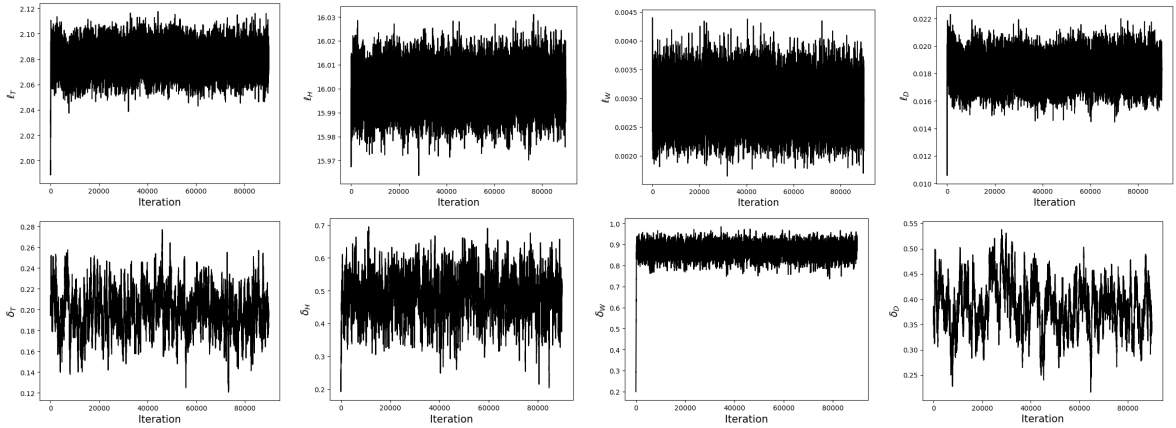

**Figure 4.** Results of learning $\ell_T$, $\ell_H$, $\ell_W$, $\ell_D$, given the multivariate training data $\mathbf{D}_2$, using our non-stationary model. In this model, the kernel hyperparameter $\ell_m$ is modelled as a random function generated by a distinct (stationary) GP, the correlation function of which is kernel-parametrised by a stationary kernel with hyperparameter $\delta_m$, $m \in \{T, H, W, D\}$. Top row, from the left: traces of $\ell_T$, $\ell_H$, $\ell_W$, $\ell_D$. Bottom row, from the left: traces of $\delta_T$, $\delta_H$, $\delta_W$, $\delta_D$.

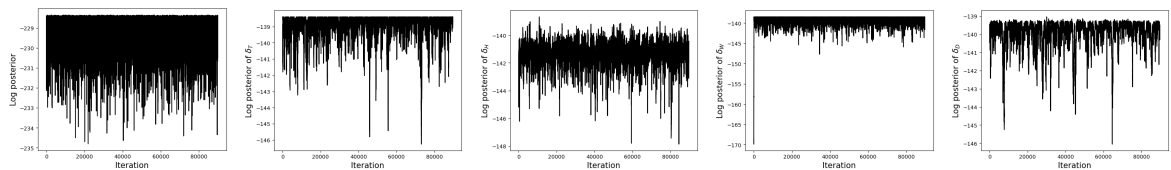

**Figure 5.** From left: trace of logarithm of the joint posterior of $\ell_T$, $\ell_H$, $\ell_W$, $\ell_D$ given training data $\mathbf{D}_2$ in our non-stationary model, and traces of logarithm of posterior of $\delta_T$, $\delta_H$, $\delta_W$, $\delta_D$ respectively, given relevant (dynamically-varying) lookback data.

**Appendix B. Results of MCMC-based inference in stationary learning.** Figure 8 displays traces of learnt $\ell_T$, $\ell_H$, $\ell_W$, $\ell_D$ using the stationary model given $\mathbf{D}_2$ and $\mathbf{D}_{2'}$. Both chains are run for 200000 runs. Normal proposal densities are used to propose values of the hyperparameters during inference. Normal priors are used on all unknown hyperparameters.

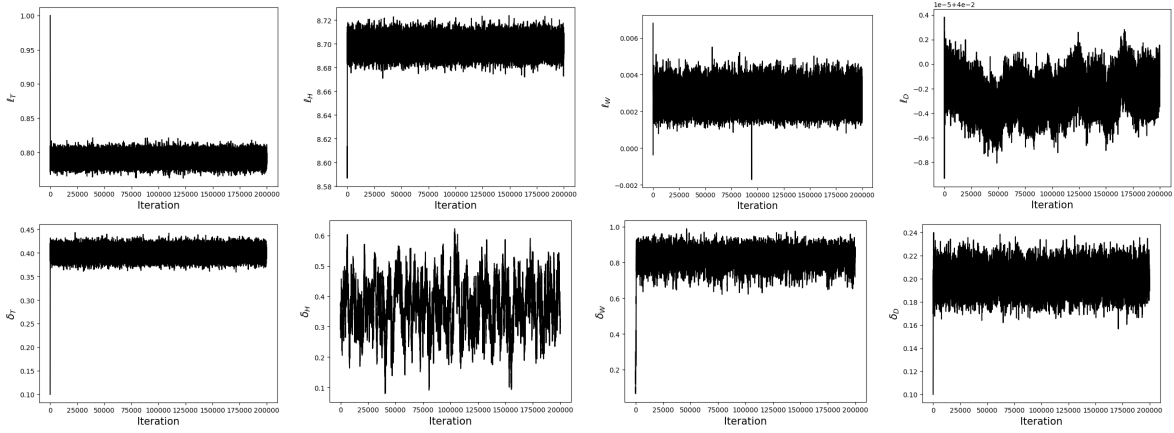

**Figure 6.** *Results of learning given the multivariate training data $\mathbf{D}_{2'}$, using our non-stationary model. Top row, from the left: traces of $\ell_T$, $\ell_H$, $\ell_W$, $\ell_D$. Bottom row, from the left: traces of $\delta_T$, $\delta_H$, $\delta_W$, $\delta_D$.*

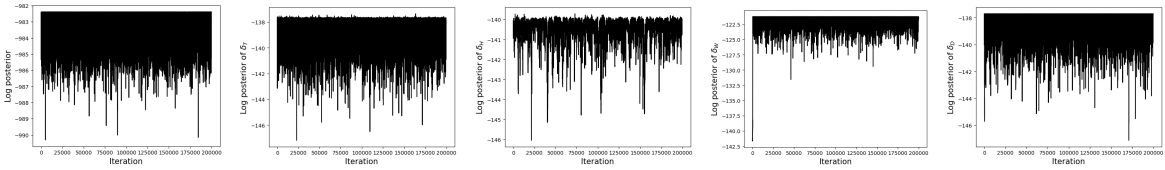

**Figure 7.** *From left: trace of logarithm of the joint posterior of $\ell_T$, $\ell_H$, $\ell_W$, $\ell_D$ given training data $\mathbf{D}_{2'}$ in the non-stationary model that we use; traces of logarithm of posterior of $\delta_T$, $\delta_H$, $\delta_W$, $\delta_D$ respectively, given relevant (dynamically-varying) lookback data.*

Variances of these Normal priors, as well as the Normal proposals are chosen experimentally.

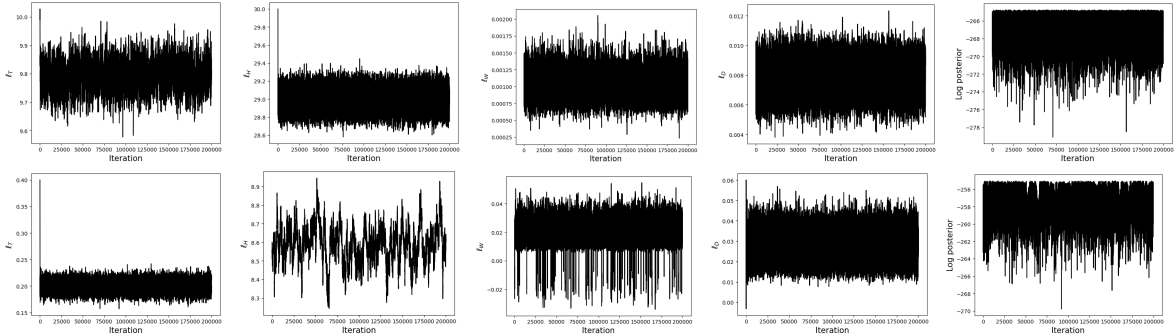

**Figure 8.** *From left: Results of learning $\ell_T$, $\ell_H$, $\ell_W$, $\ell_D$ and joint log posterior, given the training data $\mathbf{D}_2$ (top row) and $\mathbf{D}_{2'}$ (bottom row), using a stationary GP.*

**Acknowledgements.** GR was funded by an EPSRC DTP studentship when in Brunel University London, where she had started this work.